\documentclass[11pt]{article}
\usepackage[preprint]{acl}

\usepackage{times}
\usepackage{latexsym}
\usepackage[T1]{fontenc}
\usepackage[utf8]{inputenc}
\usepackage{microtype}
\usepackage{inconsolata}
\usepackage{graphicx}
\usepackage{booktabs}
\usepackage{amsmath}
\usepackage{amssymb}
\usepackage{multirow}

\title{Stuck on ``A'': Diagnosing and Repairing Interface Injury\\
in Attention-to-KDA Linearization of a 0.6B Language Model}

\author{Ronglong Bao\thanks{~~Code, weights and recipes: \url{https://github.com/Sisyphbaous-DT-Project/open-qingyi}} \\
  DT-Project \\
  \texttt{islonglongy@qq.com}}

\begin{document}
\maketitle

\begin{abstract}
We convert 21 of 28 full-attention layers of Qwen3-0.6B-Base into KDA
(Kimi Delta Attention) linear-attention layers on a single consumer-grade
GPU budget, and ask a simple question: what exactly does the conversion
break? After surgery, hidden-state alignment and end-to-end KL
distillation drive the student close to its teacher in perplexity, yet
multiple-choice accuracy stays near random chance (25--29\% vs.\ the
teacher's 50.6\% on C-Eval). Using a four-permutation diagnostic that
rotates answer options while holding content fixed, we show the model
\emph{sticks to option labels} (predicting ``A'' 81\% of the time;
106/161 questions keep the same label under all four rotations) rather
than following answer content---an \emph{interface injury} that standard
distillation metrics cannot see. A 1,000-step format-targeted
completion-only KL stage repairs the interface (+12.48 points on C-Eval,
label-stickiness roughly halved), after which persona SFT and one round
of on-policy DPO preserve benchmark scores within noise. We release
code, weights, recipes, and the full audit trail, and distill the
engineering lessons---including an FP32-master failure mode in which
bf16 optimizer updates are silently swallowed---that made convergence
possible at this budget.
\end{abstract}

\section{Introduction}

Linear attention replaces the $O(n)$ KV cache of full attention with an
$O(1)$ recurrent state, offering structural advantages for long-context
inference and edge deployment. KDA (Kimi Delta Attention), used in
Moonshot's Kimi Linear series, adds per-channel forget gating and a beta
write gate to the delta rule \citep{yang2024gated,kimilinear}. Converting
a pretrained full-attention model into such an architecture
(``linearization'') is attractive because it reuses the knowledge already
stored in the weights, but the conversion is lossy and its failure modes
are poorly documented: public records contain both successful recipes
\citep{kostelec2026gendistill} and reports of divergence
\citep{chen2026halo}.

We study linearization at the smallest practical scale---Qwen3-0.6B-Base,
28 layers, with 21 layers replaced by KDA and 7 GQA layers
retained---under a budget constraint: one rented 32\,GB GPU, tens of
millions of tokens rather than billions. Our central question is
diagnostic: \emph{what does the conversion actually damage?} The naive
reading of standard metrics is misleading twice over. First, layer-wise
hidden-state alignment reduces overall CE from 9.48 to 4.13 while
C-Eval accuracy remains at the 25\% random baseline. Second, 7,000 steps
of end-to-end forward-KL distillation close the validation CE gap to
+0.128 nats, yet accuracy rises only to 28.8\%. Perplexity, in short,
can lie \citep{kostelec2026gendistill}.

To find out where the ability went, we design a \emph{four-permutation
diagnostic}: 161 clean multiple-choice questions are built from the
training corpus itself (zero overlap with any benchmark), and each
question is evaluated four times with its options cyclically rotated.
A model that follows answer \emph{content} should track the correct
option; a model that has lost the knowledge should fail uniformly. Our
converted model does neither: it predicts ``A'' 81.06\% of the time and
keeps the same label across all four rotations on 106/161 questions,
while its average margin on the correct option is \emph{negative}
($-0.117$ nats). Uniform forgetting alone cannot explain this pattern:
the student has lost the \emph{interface} that maps knowledge onto
option labels.

A 1,000-step repair stage (Stage~3b) with completion-only KL on
teacher-verified MCQ permutations, poetry, and translation QA lifts
C-Eval from 28.8\% to 41.3\%, halves label-stickiness, and turns the
correct-option margin positive---at a measurable cost in general
distribution fit (valid KL $0.160 \to 0.196$). Persona alignment (SFT,
identity booster, one round of on-policy DPO) then proceeds without
catastrophic forgetting under 15 benchmark guard points, yielding a
final 41.83\% C-Eval model whose persona is baked into the weights.

\paragraph{Contributions.} (i) A four-permutation diagnostic that
separates \emph{interface injury} from knowledge loss in converted
models, applicable to any linearization or compression pipeline;
(ii) a complete, budget-constrained Qwen3$\to$KDA conversion recipe with
every failure documented (false teacher-init, gate ablation, bf16
update-swallowing, cache bugs); (iii) evidence that persona alignment
can survive architecture conversion essentially for free;
(iv) open release of code, weights, recipes, and audit trails.

\section{Related Work}

\paragraph{Linear attention and delta rules.}
Linear attention replaces softmax attention with recurrent state updates
\citep{katharopoulos2020transformers}. The delta rule and its gated
variants \citep{yang2024gated} underpin KDA as used in the Kimi Linear
series \citep{kimilinear}; we port the reference implementation with the
\texttt{fla} kernels \citep{fla}.

\paragraph{Linearization of pretrained models.}
HALO \citep{chen2026halo} converts Qwen3-series Transformers into hybrid
RNN-attention models with 2.3B tokens (attention weight transfer,
three-stage distillation, attention-layer selection), and reports that
Qwen3$\to$KDA conversion under its Appendix-B configuration diverges at
Stage~2 (gradient norm $\to$ inf, unresponsive to lowered learning
rates). GenDistill \citep{kostelec2026gendistill} converts Qwen3-0.6B
into a Hybrid-KDA student via layer-wise alignment, end-to-end KL, and
completion-only KD, ablating six design axes. Its headline finding is
that log-likelihood scoring systematically hides generation-quality
gaps: in its 7B motivating example, a student within 0.2 points of its
teacher under log-likelihood trails by 20.8 points under autoregressive
generation. Notably, its Appendix~N already observes a
\emph{multiple-choice positional collapse} in generation: an SFT
student picks ``A'' on 79.5\% of HellaSwag questions (vs.\ 54\% for its
KD student), with conditional accuracy on non-A options collapsing.
Our pipeline belongs to the same recipe family---our unfreezing scope
is motivated by GenDistill's MLP-freezing ablation---implemented
independently under a consumer budget. The complement is this: where
GenDistill contrasts scoring \emph{protocols} and notes positional
collapse as one of four SFT-specific generation behaviors, we make
option-label sticking the central object---a controlled four-permutation
diagnostic \emph{inside} the log-likelihood protocol that quantifies
label-following vs.\ content-following, attributes the failure to
missing format supervision, and verifies repair with before/after
re-measurement.

\paragraph{Knowledge distillation.}
We use forward KL between teacher and student distributions
\citep{hinton2015distilling}, layer-wise hidden-state MSE for Stage~2,
and completion-only masked KL for Stage~3b. Our temperature ablation
(Section~\ref{sec:temp}) relates to standard analyses of soft targets.

\paragraph{Persona alignment and preference optimization.}
Identity injection via SFT concentration control and one round of
on-policy DPO \citep{rafailov2023dpo} extends the conversion pipeline to
persona-level attributes.

\section{Conversion Pipeline}
\label{sec:pipeline}

\subsection{Hybrid Layout and Surgery}

From Qwen3-0.6B-Base \citep{qwen3}, every 4th layer keeps its native GQA
(layers 3, 7, 11, 15, 19, 23, 27, 0-indexed; the final layer is always
full attention); the remaining 21 layers are replaced with KDA (16
heads, head-dim 128, causal short conv $k{=}4$ + SiLU, q/k L2-norm,
low-rank forget gate, beta write gate, gated RMSNorm, NoPE). The KDA
implementation matches the Kimi Linear reference to $\sim$$10^{-6}$
numerical error against \texttt{fla}'s \texttt{chunk\_kda}. The hybrid
keeps a linearly growing KV cache through the 7 GQA layers; the
$O(1)$ recurrent state covers 75\% of layers.

\paragraph{KDA recurrence.}
Each KDA layer keeps a matrix-valued state
$S_t \in \mathbb{R}^{d_v \times d_k}$ per head, updated by a
per-channel gated delta rule \citep{yang2024gated,kimilinear}:
\begin{equation}
S_t = S_{t-1}\,\mathrm{Diag}(\alpha_t)\,\bigl(I - \beta_t k_t k_t^{\top}\bigr) + \beta_t\, v_t k_t^{\top},
\label{eq:kda}
\end{equation}
with output $o_t = S_t\, q_t$, where $k_t, q_t \in \mathbb{R}^{d_k}$,
$v_t, o_t \in \mathbb{R}^{d_v}$, $\alpha_t \in (0,1)^{d_k}$ is the
per-channel forget gate (median half-life $\approx$6.3 tokens under our
g6 initialization) and $\beta_t \in (0,1)$ the write gate. We use the
transposed state convention of the Kimi Linear report. Decoding state
is $O(d_k d_v)$ per layer---constant in context length---while training
runs in the parallel chunkwise form (\texttt{chunk\_kda}).

\paragraph{Gate initialization is decisive (six-scheme ablation, g1--g7).}
Judged by overall CE on a fresh held-out ruler (v2-default $=16.60$,
uniform $=11.93$, teacher $=3.01$), near-total retention explodes the
recurrent state while near-total forgetting scores \emph{worse than
random}. Our adopted setting g6 (dt\_bias $=0$, decay scale
$\exp(A_{\log}) \in (0.03, 0.3)$) reaches 9.48.

\paragraph{False teacher-init (a cautionary tale).}
Two surgery generations (v1/v2) claimed Q/K/V/O projection transplant;
forensic checks later showed v1's ``transplant'' never happened
(cosine $-0.00$ to teacher, $+0.69$ to random init) and v2's was
functionally orthogonal to the teacher (mean cosine $0.007$). v3
abandons projection transplant entirely; functional probe cosine rises
to $0.434$.

\subsection{Stage 2: Layer-wise Alignment}

Each KDA layer receives its corresponding teacher hidden state; the
teacher is frozen, errors are detached across layers, and only the KDA
parameters (193.6M) train (lr $10^{-4}{\to}10^{-5}$, 8{,}192
tokens/step). Overall CE falls from 9.48 to a best of 4.125
(KL 1.33)---\emph{while C-Eval stays at 25.5\%}. This is the cleanest
single instance of the perplexity--ability decoupling at the heart of
this paper: hidden-trajectory alignment and task ability decouple
completely.

\subsection{Stage 3a: End-to-End KL Distillation}
\label{sec:stage3a}

The objective is forward KL between teacher and student next-token
distributions over pretraining text, normalized by the number of
effective tokens, with no CE term:
\begin{equation}
\mathcal{L}_{\mathrm{KL}} = \frac{1}{N_{\mathrm{tok}}}\sum_{t} D_{\mathrm{KL}}\bigl(p_T(\cdot \mid x_{<t}) \,\|\, p_S(\cdot \mid x_{<t})\bigr),
\label{eq:kl}
\end{equation}
where $p \propto \exp(z/T)$, with the standard $T^2$ gradient rescaling
when $T \neq 1$ \citep{hinton2015distilling}. Gradients flow through
the full forward graph into all 21 KDA layers. Trainable scope: KDA +
all MLPs + all LayerNorms (457.9M); embeddings, tied lm\_head, and the
7 attention layers stay frozen, following GenDistill's MLP-freezing
ablation, which reports task-dependent but overall adverse effects on
knowledge transfer. We bypass materializing full logits
(2.32\,GiB per bf16 copy) and compute chunked fp32 KL (gradient error
$\le 2.4{\times}10^{-7}$ vs.\ dense). 7{,}000 steps at 8{,}192
tokens/step (57.3M tokens) yield valid CE 2.9385 vs.\ teacher 2.8108
(gap +0.128), valid KL 0.160---and C-Eval 28.8\%.

\subsection{Temperature Ablation}
\label{sec:temp}

Two arms resume from the same step-7000 checkpoint, data cursor,
optimizer state, and RNG: arm A at $T{=}1$, arm B at $T{=}2$, each for
1{,}000 further steps. A finishes at 28.7\% C-Eval; B at 30.5\%, holding
$\ge$30\% at two consecutive checkpoints---a directional gain in option
ranking---while B's valid CE/KL degrade (2.9385/0.1602 $\to$
2.9485/0.1702). The 1.8-point gap is not significance-tested; we treat
$T{=}2$ as a directional observation, not a causal conclusion.

\section{The Interface Injury Diagnosis}
\label{sec:diag}

\subsection{Four-Permutation Protocol}

We generate 161 clean MCQs from the training corpus itself (73 source
documents, $\le$4 questions per document; zero exact/5-gram overlap with
C-Eval \citep{huang2023ceval}, MMLU \citep{hendrycks2021mmlu}, or CMMLU
\citep{li2023cmmlu}; zero fingerprint overlap with our three-tier
evaluation rulers). Each question $i$ is scored under all four cyclic
rotations $r \in \{0,1,2,3\}$ of its options. Let
$\hat{y}_{i,r} = \arg\max_j \log p(\ell_j \mid x_{i,r})$ be the
predicted option \emph{label}, $y^{*}_{i,r}$ the correct option, and
$m_{i,r} = \log p(\ell_{y^{*}_{i,r}} \mid x_{i,r}) - \max_{j \neq y^{*}_{i,r}} \log p(\ell_j \mid x_{i,r})$
the correct-option margin under rotation $r$. We report
\begin{align}
\text{stickiness} &= \frac{1}{N}\sum_{i=1}^{N} \mathbf{1}\bigl[\hat{y}_{i,0} = \hat{y}_{i,1} = \hat{y}_{i,2} = \hat{y}_{i,3}\bigr], \label{eq:stick}\\
\text{margin} &= \frac{1}{4N}\sum_{i,r} m_{i,r}. \label{eq:margin}
\end{align}
Three behaviors separate cleanly: \emph{content-following}
(stickiness $\to$ 0, margin $>0$), \emph{knowledge loss} (uniform
failure), and \emph{label-sticking} (high stickiness, margin
collapsing)---and the student lands squarely in the third
(Table~\ref{tab:diag}).

\begin{table}[t]
\centering
\small
\begin{tabular}{@{}lcc@{}}
\toprule
Metric (161$\times$4) & Teacher & Student-7000 \\
\midrule
Accuracy & 66.46\% & 36.18\% \\
Correct-option margin (nats) & +0.995 & \textbf{$-$0.117} \\
Predicted ``A'' share & --- & \textbf{81.06\%} \\
Same label all 4 rotations & --- & \textbf{106/161} \\
Exactly 1/4 correct (luck) & --- & 112/161 \\
\bottomrule
\end{tabular}
\caption{Four-permutation diagnosis after Stage~3a. The student
overwhelmingly emits ``A'' after ``Answer:'' and fails to track answer
content---label-sticking, not uniform forgetting.}
\label{tab:diag}
\end{table}

\subsection{What the Diagnosis Does and Does Not Show}

The protocol directly evidences severe label-stickiness: pretraining-text
KL provides almost no supervision for the MCQ format. It does
\emph{not} directly prove the student ranks answer contents correctly
without labels---we lack label-free content scoring, and knowledge
damage cannot be excluded (teacher-right/student-wrong cases persist in
poetry and translation). Our claim is therefore scoped: interface injury
is \emph{one major repairable factor}, and the repair below provides
strong supporting evidence.

\section{Stage 3b: Format-Targeted Repair}
\label{sec:stage3b}

Data: 6{,}250 teacher-verified MCQs (teacher correct under all four
rotations, min margin $\ge$0.25 nats) $\times$ 4 permutations (A/B/C/D
exactly balanced) + 3{,}000 poetry QA + 3{,}000 translation QA; zero
overlap with all benchmarks and rulers. The loss is completion-only KL:
the prompt contributes no direct loss, and no question straddles a pack
boundary:
\begin{equation}
\mathcal{L}_{\mathrm{comp}} = \frac{1}{\sum_t m_t} \sum_t m_t\, D_{\mathrm{KL}}\!\Bigl(p_T \,\|\, p_S\Bigr)_t,
\label{eq:comp}
\end{equation}
where the per-token KL is conditioned on $x_{<t}$ as in
Eq.~\eqref{eq:kl}, and
$m_t = \mathbf{1}\bigl[t \in \text{completion} \cup \{\mathrm{EOS}\}\bigr]$.
Training: $T{=}2$, 1{,}000 steps (8.2M packed tokens) from step-7000.

\begin{table}[t]
\centering
\small
\begin{tabular}{@{}lcc@{}}
\toprule
 & Stage 3a & +Stage 3b \\
\midrule
C-Eval & 28.83\% & \textbf{41.31\%} \\
4-perm accuracy & 36.18\% & 53.73\% \\
All-4-correct questions & 7/161 & 47/161 \\
Correct-option margin & $-$0.117 & +0.233 \\
Predicted ``A'' share & 81.06\% & 58.5\% \\
Same label 4 rotations & 106/161 & 47/161 \\
Valid CE / KL & 2.939 / 0.160 & 2.976 / 0.196 \\
\bottomrule
\end{tabular}
\caption{Stage~3b repair. The interface improves substantially
(``substantially repaired, not cured'': ``A'' share remains far above
the 25\% balance), at a measurable cost in general distribution fit.}
\label{tab:stage3b}
\end{table}

Table~\ref{tab:stage3b} summarizes the repair: +12.48 C-Eval points,
label-stickiness roughly halved, margin turned positive. We read this as
strong evidence that \emph{most} benchmark-relevant knowledge survived
conversion and was gated behind a broken interface---not as proof of
zero knowledge loss.

\section{Persona Alignment Survives Conversion}
\label{sec:persona}

Post-repair, we inject a persona (``Qingyi'', a digital-twin assistant)
entirely at the weight level: two-epoch SFT (7{,}768 steps; lr $1.5{\times}10^{-5}$
then $10^{-5}$), a 300-step identity booster (55\% identity QA +
protective chat), and one 100-step round of on-policy DPO
\citep{rafailov2023dpo} (545 pairs, $\beta{=}0.1$):
\begin{equation}
\mathcal{L}_{\mathrm{DPO}} = -\log \sigma\bigl(\beta\,(r_w - r_l)\bigr),
\label{eq:dpo}
\end{equation}
with $r_y = \log \frac{\pi_\theta(y \mid x)}{\pi_{\mathrm{ref}}(y \mid x)}$.

\paragraph{No catastrophic forgetting.}
Fifteen rolling and terminal C-Eval evaluations were conducted during
SFT, and the booster and DPO stages were scored separately at their
acceptance gates: an early dip to 38.63\% ($-2.68$pt) recovers, and
scores stabilize at 39--41\% (final 41.83\% $\pm$ 1.33 vs.\ 41.31\%
pre-SFT---inside the noise band).

\paragraph{Concentration $\times$ steps $\times$ protection.}
Identity facts bind at step-300 of the booster; step-400 overflows
(denying H$_2$O, greedy loops in chit-chat). There is no universal
concentration threshold: 26\% succeeded in an earlier generation, 55\%
succeeds here at 300 steps---the safe window is a three-variable
interaction.

\paragraph{One DPO round is the sweet spot.}
Held-out reward accuracy $0.625 \to 0.792$; a second on-policy round
only breeds repetition and is discarded wholesale.

\section{Engineering Lessons}
\label{sec:lessons}

\paragraph{bf16 swallows small updates.}
With bf16 parameters and bf16 Adam moments, a norm weight initialized at
$1.0$ remains \emph{exactly} $1.0$ after 1{,}000 steps at lr $10^{-4}$
(sub-ULP updates). Stages 2/3 use FP32 master weights and moments. Any
distillation run showing healthy gradient norms but frozen weights
should check this first---we note this matches HALO's
``unresponsive to lowered lr'' symptom, though we claim no causal proof.

\paragraph{New layers need cache parity from day one.}
Our KDA layer initially accepted but ignored the incremental cache,
collapsing generation into garbage from the second token---and several
``persona not bound'' readings were artifacts of this bug. The fix
(\texttt{HybridKDACache}: per-layer recurrent state + conv windows)
removes the collapse at 1.7$\times$ speedup, no retraining.

\paragraph{Process hardening.}
Canonical-init SHA-256 locking at the training entry; three-tier
evaluation rulers with document-fingerprint isolation (tune/valid/release;
release never opened); resume validation of 11 hyperparameters;
cursor-exact data replay. Each rule was earned by a concrete accident.

\section{Results}
\label{sec:results}

\begin{table}[t]
\centering
\small
\begin{tabular}{@{}llc@{}}
\toprule
Stage & Checkpoint & C-Eval \\
\midrule
Base & Qwen3-0.6B-Base & 50.52 / 50.59\% \\
v3 & surgery (untrained) & --- (CE 9.48) \\
v3 & Stage 2 best & 25.48\% \\
v3 & Stage 3a step-7000 & 28.83\% \\
v3 & Stage 3b step-8000 & \textbf{41.31\%} \\
v3 & SFT step-7768 & 40.34\% \\
v3 & booster step-300 & \textbf{42.27\%} \\
v3 & DPO step-100 (final) & \textbf{41.83\% $\pm$ 1.33} \\
\bottomrule
\end{tabular}
\caption{Full lineage (lm-evaluation-harness \citep{eval-harness},
\texttt{ceval-valid}, 0-shot). The two teacher readings are
same-protocol measurements at different times.}
\label{tab:lineage}
\end{table}

Table~\ref{tab:lineage} collects the lineage. The final model sits
8.8 points below its teacher; whether the residual gap is data diversity
(our working hypothesis) or architectural capacity is undecidable from a
single model, seed, and data scale.

\section{Limitations}
\label{sec:limits}

(i) The same \texttt{ceval-valid} served repeatedly for checkpoint
selection (15+ guard points), so the final score is not an untouched
test (winner's curse); our locked release ruler was never opened before
submission. (ii) Four-permutation evidence does not include label-free
content scoring; knowledge damage is not excluded. (iii) No final
MMLU/CMMLU, generation, long-context, or throughput benchmarks on the
final weights. (iv) Refusal behavior under sampling is unstable.
(v) Single model/seed/scale; no KDA CPU fallback (Triton kernels).
(vi) The temperature ablation lacks significance testing.

\section{Conclusion}

Converting a 0.6B full-attention model to a KDA hybrid on a consumer
budget is feasible, but the standard metric stack---perplexity, KL,
hidden-state MSE---systematically misreports the damage. A
four-permutation diagnostic reveals a major repairable injury: the
interface from knowledge to option labels. A short format-targeted
stage repairs it, and persona alignment then costs essentially nothing.
We release everything: code, weights, recipes, and the accidents.

\section*{Acknowledgments}

Compute: one rented 32\,GB GPU; local RTX 4070 Laptop for v1. We thank
the authors of GenDistill and HALO for public recipes and failure
records.

\bibliography{custom}

\end{document}